%% file: root.tex
\documentclass[a4paper,conference]{IEEEtran}
\ifCLASSINFOpdf
  \usepackage[pdftex]{graphicx}
  \graphicspath{{./fig/}}
  \DeclareGraphicsExtensions{.pdf,.jpeg,.png}
\else
\fi
\usepackage{amsmath}
\usepackage{multirow}

\usepackage{subfig}
\usepackage{url}
\usepackage{hyperref}

\begin{document}
%
\title{F2DNet: Fast Focal Detection Network \\ for Pedestrian Detection}




%
\author{
    \IEEEauthorblockN{
        Abdul Hannan Khan\IEEEauthorrefmark{1}\IEEEauthorrefmark{2},
        Mohsin Munir\IEEEauthorrefmark{2},
        Ludger van Elst\IEEEauthorrefmark{2} and
        Andreas Dengel\IEEEauthorrefmark{1}\IEEEauthorrefmark{2},
    }
    \IEEEauthorblockA{
        \IEEEauthorrefmark{1}Fachbereich Informatik,
        Technische Universität Kaiserslautern, \\
        67663 Kaiserslautern, Germany
    }
    \IEEEauthorblockA{
        \IEEEauthorrefmark{2}German Research Center for Artificial Intelligence (DFKI GmbH),\\ 
        67663 Kaiserslautern, Germany
    }
    \IEEEauthorblockA{
        Corresponding Author: hannan.khan@dfki.de
    }
}


\maketitle

\input{abstract.tex}


%
\IEEEpeerreviewmaketitle

\input{intro}

\input{related_work}

\input{method}

\input{experiments}

\input{Results}

\input{conclusion}


\section*{Acknowledgment}
This work was supported by the German Ministry for Economic Affairs and Climate Action (BMWK) project KI Wissen under Grant 19A20020G.




\bibliographystyle{IEEEtran}
%
\bibliography{root}

\end{document}

%% file: abstract.tex
\begin{abstract}
Two-stage detectors are state-of-the-art in object detection as well as pedestrian detection. However, the current two-stage detectors are inefficient as they do bounding box regression in multiple steps i.e. in region proposal networks and bounding box heads. Also, the anchor-based region proposal networks are computationally expensive to train. We propose F2DNet, a novel two-stage detection architecture which eliminates redundancy of current two-stage detectors by replacing the region proposal network with our focal detection network and bounding box head with our fast suppression head. We benchmark F2DNet on top pedestrian detection datasets, thoroughly compare it against the existing state-of-the-art detectors and conduct cross dataset evaluation to test the generalizability of our model to unseen data. Our F2DNet achieves 8.7\%, 2.2\%, and 6.1\% $MR^{-2}$ on City Persons, Caltech Pedestrian, and Euro City Person datasets respectively when trained on a single dataset and reaches 20.4\% and 26.2\% $MR^{-2}$ in heavy occlusion setting of Caltech Pedestrian and City Persons datasets when using progressive fine-tunning. Furthermore, F2DNet have significantly lesser inference time compared to the current state-of-the-art. Code and trained models will be available at \url{https://github.com/AbdulHannanKhan/F2DNet}.

\end{abstract}

%% file: intro.tex
\section{Introduction}

Pedestrian detection is a sub-domain of object detection where the target class is pedestrian and the rest is considered background. Pedestrian detection plays a vital role in autonomous driving as well as surveillance. In autonomous driving, one of the most important objectives is to avoid collision with pedestrians by detecting and tracking them. This objective is to be carried out in a limited resource scenario as limited computational power is available inside an autonomous vehicle due to compactness and power efficiency constraints. This requires the pedestrian detection model to be light and efficient. Also, the lesser the time model takes to process a single frame more frame per second it can process which yields better awareness of surroundings.

Region Proposal Networks were first proposed by Ross Girshick et al. \cite{fasterrcnn} to replace, slow, selective search-based region proposal generation with a faster, CNN-based network that can be trained end-to-end along with detection head. In the last decade, researchers have focused on improving two-stage detectors by proposing new detection heads \cite{mgan, betarcnn, cascadercnn} with little focus on region proposal network architecture. However, the role of region proposal networks in two-stage detectors is limited to proposing candidate regions with the purpose of objectness score produced by region proposal networks limited to proposal filtering. Also, proposed bounding boxes from region proposal network need rigorous refinement in their coordinates, for example, Cascade RCNN \cite{cascadercnn} applies three cascading heads to get refined detections.

\begin{figure}
    \centering
    \includegraphics[width=0.5\textwidth]{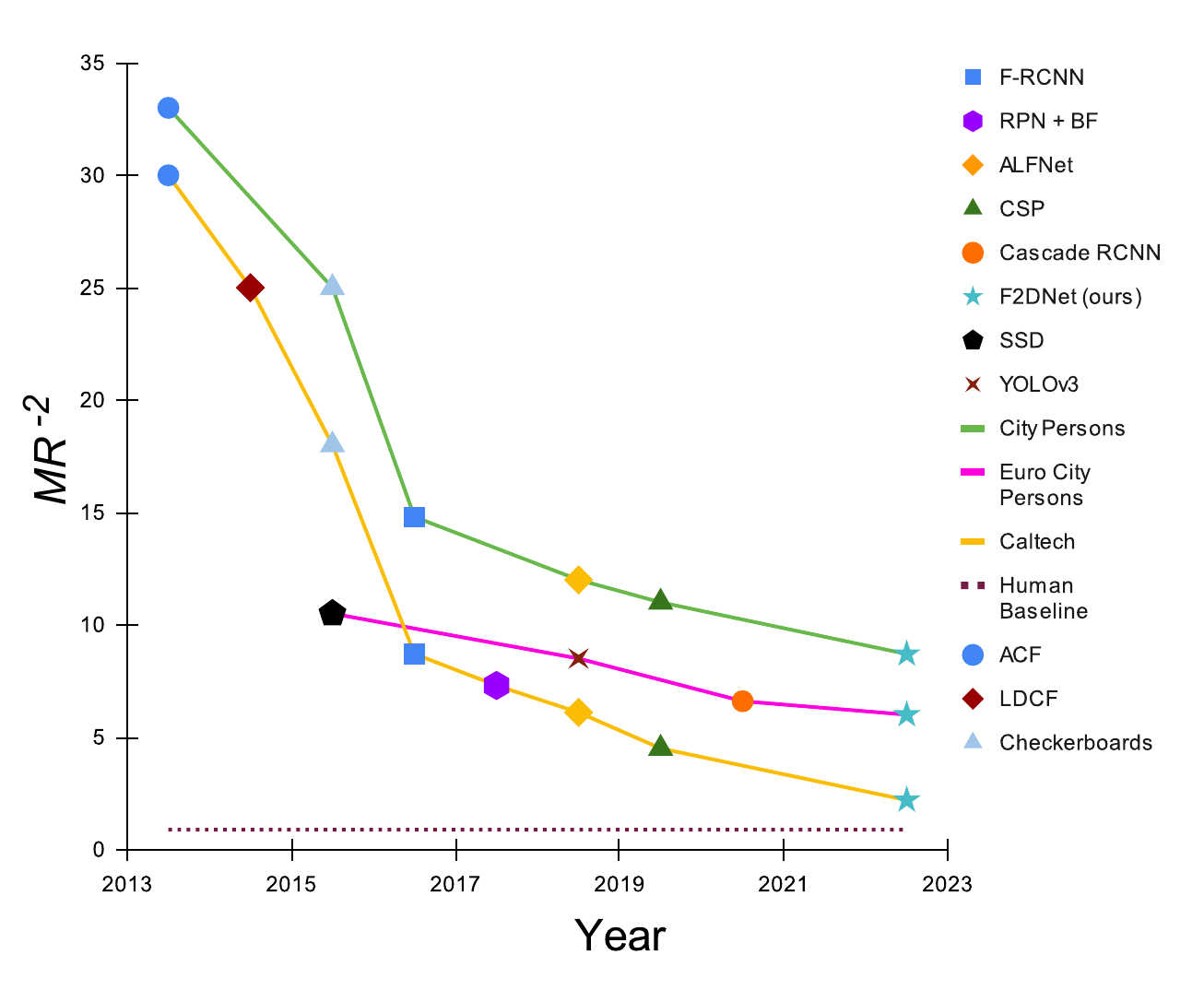}
    \caption{Evolution of pedestrian detectors over the years and their corresponding $MR^{-2}$ on Caltech Pedestrian \cite{caltech} (orange), City Persons \cite{citypersons} (green) and Euro City Persons \cite{eurocitypersons} (pink) datasets in reasonable settings.}
    \label{fig:history}
\end{figure}

Compared to two-stage detectors, single-stage detectors are efficient as they split the image into a grid and perform detection per patch eliminating region proposal network \cite{ssd, yolo, focalloss, alfnet}. However, single-stage detectors do not perform as good as two-stage detectors in terms of accuracy, this can be attributed to a class imbalance between positive and negative samples \cite{focalloss}. Other than class imbalance, since each patch does not necessarily contain a full object, classifying if a patch contains enough parts of an object is sub-optimal, as a part may belong to multiple object classes. A common attribute of both single and two-stage detectors explained above is anchors. Both kinds of detectors rely on anchors with predefined aspect ratios.

In the last few years, anchor-free object detectors were proposed \cite{fcos, cornernet}. Motivated from anchor-free approaches in object detection, anchor-free approaches were proposed for pedestrian detection as well \cite{csp, acsp, apd}. These pedestrian-specific approaches, take the idea of single-stage detector to another level by predicting classes per pixel instead of per patch. However, this is done on a downscaled feature map to be efficient and robust \cite{csp}. Unlike anchor-based approaches, center and scale-based approaches classify if each pixel is a center pixel of an object and regress the possible scale of that object. In this way, center and scale-based approaches eliminate the idea of predicting rough bounding boxes and refining them later on. Further, center and scale-based approaches use the focal loss as classification loss to deal with class imbalance \cite{csp}.

Although center and scale-based approaches have optimal design and better convergence they produce more false positives due to penalty reduced focal loss, which does not punish much the false predictions in the neighborhood of positive pixels. This problem intensifies in the case of small and heavily occluded pedestrians.

Our method is different in nature from existing single and two-stage detectors. The closest single-stage detector is anchor-free, center and scale prediction \cite{csp}. However, we only use the head from CSP \cite{csp} with different loss settings as the CSP head \cite{csp} is stronger and efficient compared to region proposal networks, also we use fast suppression head to further refine detections. Compared to two-stage detectors, we replace the region proposal network with a stronger detection network, we do not call it another region proposal network because focal detection network produces strong detection candidates compared to proposals that need further bounding box refinement and classification. Also, we replace computationally expensive traditional second stage, which predicts bounding boxes as well as classifies them, with a simple and efficient suppression head to only suppress false positives without altering bounding boxes.

Contribution of this paper is three fold;
\begin{itemize}
 
\item First, we redesign a two-stage detection architecture to remove redundant and inefficient bounding box prediction and replace region proposal network with a strong detection network, followed by a light-weight suppression head instead of multiple bounding box heads.

\item Second, we propose focal detection network as our classification and bounding box regression head, which can independently produce satisfactory results.

\item Third, we propose fast suppression head to handle false positives produced by focal detection head in small and heavily occluded settings.
\end{itemize}

\begin{figure*}[!t]
    \centering
    \includegraphics[width=0.9\textwidth]{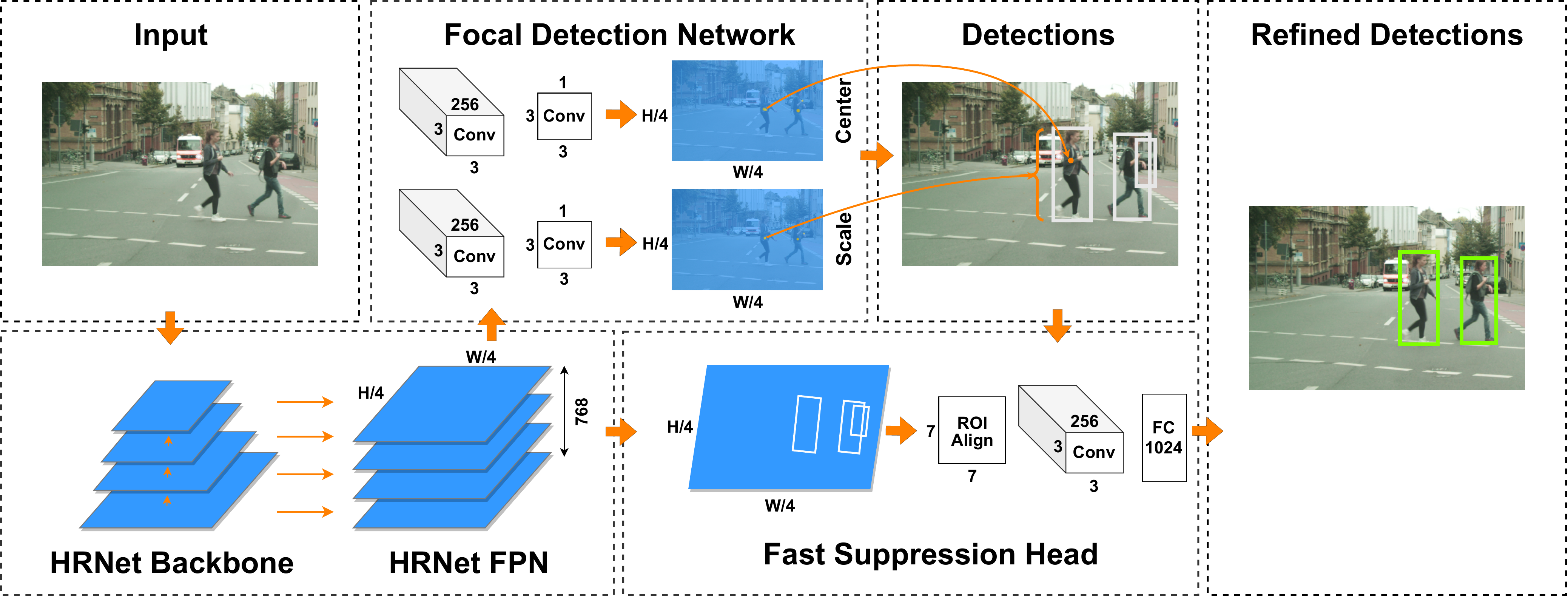}
    \caption{The network architecture of our F2DNet. The input image is passed through backbone and FPN to extract feature maps which are then passed to the focal detection network to obtain initial detections. The detected bounding boxes are then passed to the fast suppressed head along with feature maps to suppress false positives.}
    \label{fig:f2d}
\end{figure*}

%% file: related_work.tex
\section{Related Work}
Significant improvements have been made in recent years in the field of pedestrian detection using deep learning models \cite{pedestron, mgan, alfnet, csp} as shown in Fig. \ref{fig:history}. Most of the recent techniques follow general object detection workflow including a strong pre-trained backbone to extract features, an optional feature pyramid network (FPN) \cite{fpn} based feature enrichment layer, a region proposal network (RPN) \cite{fasterrcnn} in case of two-stage detectors and at the end, bounding box heads for bounding box regression and classification. Such pipelines are supported in modern object detection frameworks like mmdetection \cite{mmdetection}. Different types of pedestrian detectors have emerged in recent years which can be differentiated from each other based on how they use region proposal network and choice of bounding box heads.

\subsection{Anchor Based Pedestrian Detectors}
Region-based convolutional neural networks are two-staged object detectors, which were first proposed by Girshick et al. in \cite{rcnn} for object detection. Fast-RCNN and Faster-RCNN were proposed to improve the processing time of RCNN by using ROI pooling on features maps instead of raw image and CNN-based region proposal network respectively \cite{fastrcnn, fasterrcnn}. Mask Guided Attention Network incorporates additional visibility information of the object to handle occlusions better \cite{mgan}. Cascade R-CNN proposed by Cai and Vasconcelos in \cite{cascadercnn} uses multiple bounding box heads to refine detections in cascading manner. Another anchor-based but single-stage pedestrian detector is ALFNet, which is based on Single Shot Multibox Detector (SSD) \cite{alfnet, ssd}. RetinaNet is yet another single-stage detector, which is similar to SSD \cite{ssd} but introduces focal loss to handle foreground and background class imbalance \cite{retinanet}.

\subsection{Anchor Free Pedestrian Detectors}
Anchor-free pedestrian detectors are pedestrian-specific object detectors that do not use anchors or region proposal networks. Instead, they predict bounding box and class per pixel on down-scaled feature maps because performing detection per pixel on original resolution is costly. CornerNet \cite{cornernet} uses CNN based approach to predict paired keypoint heatmaps i.e. one heatmap for each top-left and bottom-right corner. Fully convolutional single-stage object detection network, FCOS was proposed in \cite{fcos}, which adopts classification and bounding box prediction of R-CNN heads to pixel-wise fashion with bounding box predictions being pixel distances from object center, which is calculated using centeredness predicted per pixel by FCOS. Center and Scale Prediction CSP \cite{csp} proposed for pedestrian detection uses a similar approach but instead predicts center heatmap, scale map and reconstructs the bounding boxes using the center and scale \cite{csp}. ACSP \cite{acsp} uses switchable normalization for better convergence on different batch sizes and
uses full resolution for training to improve recall. APD \cite{apd} tries to handle crowded pedestrians by additionally predicting density and diversity. BGCNet replaces normal convolutions with box-guided convolution for center heatmap subnet to incorporate predicted scale and offset information in center heatmap prediction \cite{bgcnet}. 

\subsection{ViT Based Methods}
Vision transformer (ViT) is the adaptation of transformers in the domain of computer vision. DETR \cite{detr} and DETR for pedestrian detection \cite{ped_detr} use ViT based bounding box heads to provide a wider receptive field. Recently proposed soft teacher based approach for object detection \cite{softteacher} uses SWin transformer \cite{swint} based backbone and semi-supervised training to produce promising results.

%% file: method.tex
\section{F2DNET: Fast Focal Detection Network}
Current two-stage object detection architectures employ a weak region proposal network followed by strong bounding box heads. We take a different approach and use a strong detection head succeeded by a light suppression head. In this way, the detection head focuses on precise localization and high classification recall while the suppression head takes care of false positives. In short, our two-stage detection architecture attains high efficiency by eliminating the repetition contained in current two-stage architectures.

In this section, we explain the architecture of our fast focal detection network in detail and argue about our design choices. First, we elaborate on the feature extraction process followed by the detailed architecture of F2DNet and conclude the section by explaining the detection formation strategy. Fig. \ref{fig:f2d} shows complete architecture of our model.

\subsection{Feature Extraction}
To predict precise location and size high-resolution features are required which contain semantic and position information. Aggressive down and upscaling can result in loss of this vital information \cite{pedestron}. Therefore, we use the HRNetW32v2 backbone \cite{hrnet} for feature extraction as it extracts high-resolution features from images.
To obtain feature maps of a single scale, we take feature maps from different stages of backbone, upscale them to $(h/4, w/4)$ using bilinear interpolation and apply convolution operations. In this way, the model stays light on memory as interpolation operation has no memory cost but is effective as succeeding convolution operations provide necessary learnable parameters.

\subsection{Focal Detection Network}
The architecture of the focal detection network is based on the idea of center and scale map prediction which eliminates explicit modeling of bounding boxes for detection \cite{csp}. Our approach is somewhat similar to that proposed in \cite{csp} however, we use different loss settings to fine-tune the architecture for better convergence and precise localization.

Center loss for focal detection network can be formulated as: 

\begin{equation}
    \label{eq:focal_loss}
    L_{center} = \frac{1}{K} \sum_{i} \sum_{j} \alpha_{ij}\, CE(p_{ij}, y_{ij}),
\end{equation}

where

\begin{equation}
\begin{aligned}
    \label{eq:fl_gau}
    CE(p_{ij}, y_{ij}) = &\begin{cases}
          - \log(p_{ij}) \quad &\text{if} \, y_{ij} = 1 \\
          - \log(1 - p_{ij}) \quad &\text{otherwise},\\
     \end{cases}
     \\
        \alpha_{ij} = &\begin{cases}
          (1 - p_{ij})^{\gamma} \quad &\text{if} \, y_{ij} = 1 \\
          p_{ij}^{\gamma} (1 - M_{ij})^{\beta} \quad &\text{otherwise}.\\
     \end{cases}
\end{aligned}
\end{equation}

 In equation above, $p_{ij}$ and $y_{ij}$ are predicted center probability and ground truth label respectively. $CE(p_{ij}, y_{ij})$ represents cross entropy loss with $\alpha_{ij}$ being weight at each location $(i, j)$. $M_{ij}$ represents gaussian based penalty reduction for surrounding pixels of true centers as designation of exact center brings difficulty in training \cite{csp}. The $p_{ij}^{\gamma}$ and $(1 -p_{ij})^{\gamma}$ terms define focus weight based on prediction confidence i.e. it reduces contribution of easy examples to the loss and helps optimizer to focus on hard examples. The $(1 - M_{ij})^{\beta}$ term reduces loss for false positives closer to true centers. We found $\gamma = 2$ and $\beta = 4$ to work best in our experiments.

In \cite{fastrcnn} Smooth L1 loss is recommended for regression as it is robust to outliers. The Smooth L1 loss reduces penalty when the distance between predicted and actual height is small, which helps in better convergence. However, since we use $\log$ of height instead of actual height value it can cause smaller detections and ultimately result in false positives due to insufficient IoU. Therefore, we use Vanilla L1 Loss as regression loss to make height predictions more accurate.

We define loss for the focal detection head as:

\begin{equation}
    \label{eq:loss}
    L_{FDN} = \lambda_r\, L_{reg} + \lambda_c\, L_{cls} + \lambda_o\, L_{off}
\end{equation}

Where $\lambda_r$, $\lambda_c$ and $\lambda_o$ represent weights for regression, classification and offset loss respectively. We experimentally found $\lambda_r = 0.05$, $\lambda_c = 0.01$ and $\lambda_o = 0.1$ help model converge better than other weight settings.

\subsection{Fast Suppression Head}
Since, the focal detection network uses penalty-reduced focal loss as a center loss, false positives in the neighborhood of positive centers are not punished sufficiently. While most of these false positives are suppressed by Non-Maximum Suppression (NMS), it still needs another suppression step to suppress the rests i.e. where IoU with positive predictions is lower than $0.5$. Therefore, we propose a simple and fast suppression head to further refine the detections. The fast suppression head comprises of ROI Align layer followed by convolutional and dense layers as shown in Fig. \ref{fig:f2d}. We train the fast suppression head in detached settings, i.e. the gradients from the fast suppression head do not flow back to feature maps or detection head. In this way, a simple, light yet effective suppression head is achieved. We use binary cross entropy as loss for our fast suppression head.

\begin{figure}
    \centering
    \includegraphics[width=0.5\textwidth]{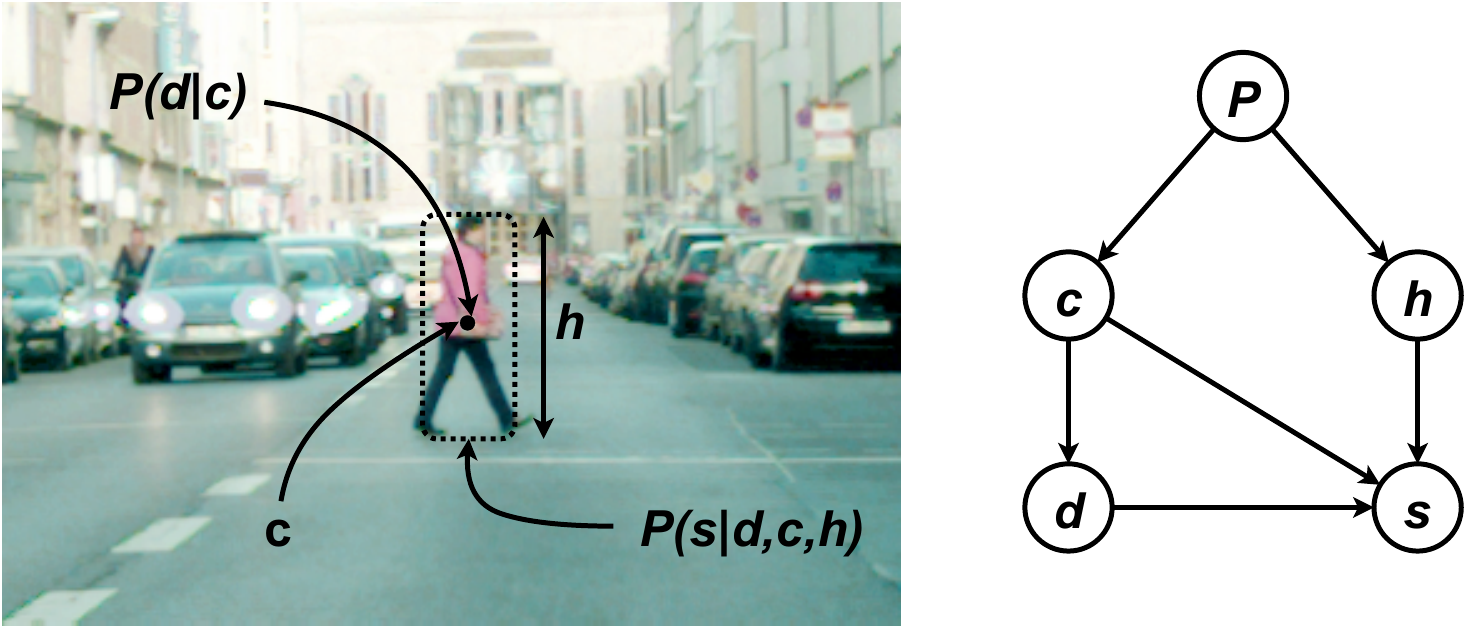}
    \caption{(Left): Representation of pedestrian showing: the center $c$, height $h$, prediction from focal detection network $P(d|c)$ and prediction from fast suppression head $P(s|d,c,h)$. (Right): graphical representation of our pedestrian detection generative model where $P$ represents pedestrian.}
    \label{fig:generative_model}
\end{figure}

\subsection{Pedestrian Detection}
Each prediction gets one score from the focal detection network and another from the fast suppression head. We eliminate thresholding hyperparameter by combining both scores using the generative model shown in Fig. \ref{fig:generative_model}. We are particularly interested in an event where pedestrian is detected an not suppressed i.e. $P(\neg s,d|c,h)$. The detection model is derived from joint probability distribution of $P(s,d,c,h)$ and represented by following relation:

\begin{equation}
    \label{eq:score}
    \begin{aligned}
    P(\neg s,d|c,h) = &P(\neg s|d,c,h) P(d|c)
    \end{aligned}
\end{equation}

\begin{table}[!t]
    \renewcommand{\arraystretch}{1.3}
    \caption{Pedestrian detection datasets summary.}
    \label{tab:datasets}
    \centering
    \begin{tabular}{l|c|c|c|c}
         \hline
         Dataset & Images & Pedestrians & Density & Resolution \\
         \hline
         Caltech Pedestrians & 42,782 & 13,674 & 0.32 & 640 $\times$ 480 \\
         \hline
         City Persons & 2,975 & 19,238 & 6.47 & 2048 $\times$ 1024 \\
         \hline
         Euro City Persons & 21,795 & 201,323 & 9.2 & 1920 $\times$ 1024 \\
         \hline
    \end{tabular}
\end{table}

where $c$ and $h$ are the center and height of pedestrian respectively. $P(d|c)$ is the probability of a position detected as pedestrian center by focal detection head and $P(s|d,c,h)$ is the probability that given a bounding box detection it is suppressed by fast suppression head.

%% file: experiments.tex
\section{Experimental Setup}
In this section, our experimental setup is detailed, which we follow in the rest of the paper unless stated otherwise. First, we briefly go through datasets, followed by the evaluation settings in which we evaluate results on these datasets, and finally, we explain the evaluation criteria we used to compare our models with the existing state-of-the-art.

\begin{table}[!t]
    \renewcommand{\arraystretch}{1.3}
    \caption{Evaluation settings for pedestrian datasets based on height and visibility.}
    \label{tab:eval_settings}
    \centering
    \begin{tabular}{l|c|c|c|c}
        \hline
        Setting & 
        \multicolumn{2}{c}{
            \begin{tabular}{c c}
             \multicolumn{2}{c}{\textbf{City Persons, Caltech}}  \\
             \hline
             Visibility & Height 
        \end{tabular}
        } &
        \multicolumn{2}{|c}{
            \begin{tabular}{c c}
             \multicolumn{2}{c}{\textbf{Euro City Persons}}  \\
             \hline
             Visibility & Height 
        \end{tabular}
        }
        \\
         \hline
         Reasonable & [0.65, $\infty$] & [50, $\infty$]& [0.6, $\infty$] & [40, $\infty$]\\
         \hline
         Small & [0.65, $\infty$] & [50, 75] & [0.6, $\infty$] & [30, 60] \\
         \hline
         Heavy Occlusion & [0.2, 0.65] & [50. $\infty$] & [0.2, 0.6] & [40. $\infty$] \\
         \hline
         All & [0.2, $\infty$] & [20. $\infty$] & [0.2, $\infty$] & [20. $\infty$]\\
         \hline
    \end{tabular}
\end{table}

\subsection{Datasets}
To benchmark, our model we present results on three commonly used pedestrian detection datasets i.e. City Persons \cite{citypersons}, Euro City Persons \cite{eurocitypersons} and Caltech Pedestrian dataset \cite{caltech}. Detailed statistics of these datasets can be seen in Table \ref{tab:datasets}.

All the results presented in this paper for Caltech pedestrian dataset \cite{caltech} are based on its test set, while for City Persons \cite{citypersons} and Euro City Persons \cite{eurocitypersons} results are based on their respective validation sets, unless stated otherwise. Also, new annotations for Caltech pedestrian dataset \cite{caltech} proposed in \cite{calnew} were used.

\subsection{Evaluation Settings}
In pedestrian detection, evaluation settings define different subsets of a dataset which are used to better judge the performance of a model in different scenarios. We use evaluation settings proposed in Caltech Pedestrian \cite{caltech} and Euro City Persons \cite{eurocitypersons} datasets. Based on visibility and height of annotations these evaluation settings form four groups where each annotation can belong to more than one group. Settings followed across the paper can be seen in the Table \ref{tab:eval_settings}. It is important to note that evaluation settings are different for Euro City Persons dataset \cite{eurocitypersons} while City Persons \cite{citypersons} and Caltech Pedestrian datasets \cite{caltech} share identical evaluation settings.

\subsection{Evaluation Criteria}
We use Log-average miss rate over false positive per image or $MR^{-2}$ to compare our model against recent models as it has been suggested in pedestrian detection datasets \cite{citypersons, caltech, eurocitypersons} as well as followed by the state of the art \cite{pedestron, bgcnet, betarcnn, csp}. $MR{-2}$ is calculated by taking geometric mean of miss rates at 9 equally spaced $ffpi$ thresholds in log space i.e. $fppi \in \{10^{-2}, 10^{-1.75}, ..., 10^{0}\}$.

\subsection{Weighted Averaging}
We used the mean teacher strategy of weighted averaging for better convergence and performance, as the model obtained after the weighted averaging performs better \cite{mean_teacher, csp}. All results of our models provided in this paper are based on the evaluation of the averaged model unless stated otherwise.

\subsection{Training Details}
We used the Nvidia RTXA6000 GPU cluster to train our models. We used Distributed Data-Parallel to achieve parallel training on multiple GPUs with a manual seed. We used $2$ GPUs with $32$ and $4$ images per GPU for training model on Caltech Pedestrian \cite{caltech} and City Persons \cite{citypersons} datasets respectively. However, for training the model on Euro City Persons dataset \cite{eurocitypersons} we used $4$ GPUs with $4$ images per GPU. We used a constant learning rate throughout the training after warm-up iterations with a maximum of $80$ epochs.

%% file: results.tex
\begin{figure}[!t]
\centering
\subfloat[][]{\includegraphics[width=0.155\textwidth]{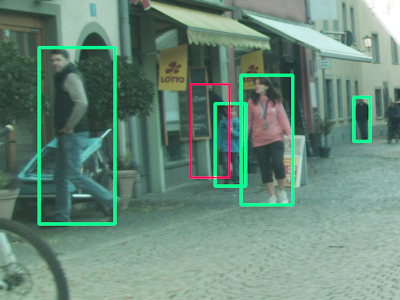} \label{fig:sota_csc1}}
\subfloat[][]{\includegraphics[width=0.155\textwidth]{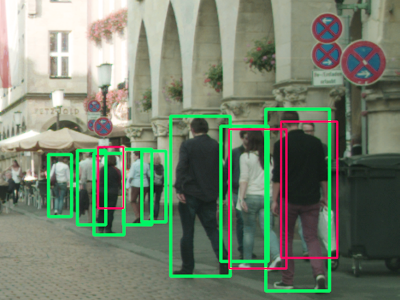} \label{fig:sota_csc2}}
\subfloat[][]{\includegraphics[width=0.155\textwidth]{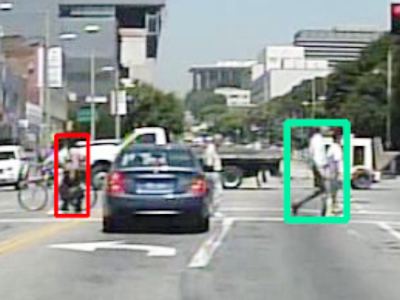} \label{fig:sota_csc3}}
\vspace{-0.6em}
\subfloat[][]{\includegraphics[width=0.155\textwidth]{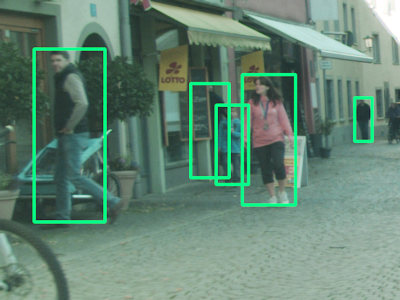} \label{fig:sota_f2d1}}
\subfloat[][]{\includegraphics[width=0.155\textwidth]{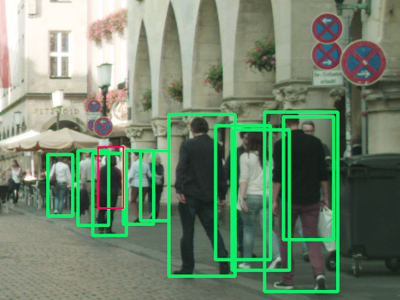} \label{fig:sota_f2d2}}
\subfloat[][]{\includegraphics[width=0.155\textwidth]{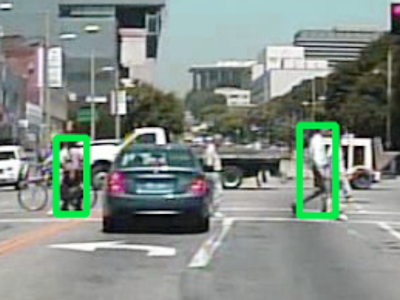} \label{fig:sota_f2d3}}
\caption{Qualitative comparison of F2DNet and Cascade R-CNN results. (a, b, c) Results of Cascade R-CNN on City Perons \cite{citypersons} and Caltech Pedestrians \cite{caltech} datasets. Bounding Boxes marked red indicate false negatives. (d, e, f) Results of F2DNet on City Persons \cite{citypersons} and Caltech Pedestrian \cite{caltech} datasets.}
\label{fig:sota_comp}
\end{figure}

\begin{table}[!t]
        \renewcommand{\arraystretch}{1.3}
        \caption{Comparison of F2DNet with the current state-of-the-art detectors based on $MR^{-2}$ and inference Time.}
        \label{tab:sota}
        \centering
        \begin{tabular}{l|c|c|c|c}
             \hline
             \multicolumn{1}{c|} {Method\textbackslash $MR^{-2}$} & Reasonable & Small & Heavy Occ.& Time \\
             \hline
             \hline
             \multicolumn{5}{c}{\textbf{City Persons} \cite{citypersons}}\\
             \hline
             \hline
             ALFNet \cite{alfnet} & 12.0 & 19.0 & 51.9 & 0.27s  \\
             \hline
            Cascade R-CNN \cite{pedestron} & 11.2 & 14.0 & 37.0 & 0.73s  \\
            \hline
            CSP \cite{csp}  & 11.0 & 16.0 & 49.3 & 0.33s  \\
             \hline
            PRNet \cite{prnet}  & 10.8 & - & 42.0 & 0.22s  \\
            \hline
            Beta R-CNN \cite{betarcnn} & 10.6 & - & 47.1 & -  \\
            \hline
            MGAN \cite{mgan}  & 10.5 & - & 39.4 & -  \\
            \hline
            Adaptive CSP \cite{acsp}  & 10.0 & - & 46.1 & -  \\
              \hline
            F2DNet no sup. (ours)  & 9.0 & 11.5 & 33.8 & 0.43s  \\
            \hline
             BGCNet \cite{bgcnet}  & 8.8 & 11.6 & 43.9 & - \\
            \hline
            F2DNet (ours) & \textbf{8.7} & \textbf{11.3} & \textbf{32.6} & 0.44s  \\
             \hline
             \hline
             \multicolumn{5}{c}{\textbf{Caltech} \cite{caltech}} \\
             \hline
             \hline
             Cascade R-CNN \cite{pedestron} & 6.2 & 7.4 & 55.3 & 0.20s  \\
             \hline
             ALFNet \cite{alfnet}  & 6.1 & 7.9 & 51.0 & 0.05s  \\
             \hline
             CSP \cite{csp}  & 5.0 & 6.8 & 46.6 & -  \\
             \hline
             Rep Loss \cite{reploss}  & 5.0 & 5.2 & 47.9 & -  \\
             \hline
              F2DNet no sup. (ours)  & 2.3 & 2.7 & \textbf{38.2} & 0.13s  \\
             \hline
             F2DNet (ours) & \textbf{2.2} & \textbf{2.5} & 38.7  & 0.14s \\
             \hline
             \hline
             \multicolumn{5}{c}{\textbf{Euro City Persons} \cite{eurocitypersons}} \\
             \hline
             \hline
             SSD \cite{eurocitypersons} &  10.5 & 20.5 & 42.0 & -  \\
             \hline
             YOLOv3 \cite{eurocitypersons} &  8.5 & 17.8 & 37.0 & -  \\
             \hline
             Faster R-CNN \cite{eurocitypersons} &  7.3 & 16.6 & 52.0 & -  \\
             \hline
             F2DNet no sup. (ours) &  7.2 & 12.8 & 31.6  & 0.40s \\
             \hline
             Cascade RCNN \cite{pedestron} &  6.6 & 13.6 & 33.3  & 0.44s \\
             \hline
             F2DNet (ours) &  \textbf{6.1} & \textbf{10.7} & \textbf{28.2}  & 0.41s \\
             \hline
        \end{tabular}
    \end{table}

\section{Experiments and Results}

In this section, we present the comparison of F2DNet to the current state-of-the-art and top-performing detectors based on $MR^{-2}$ and inference time, along with the performance gains achieved by the suppression head. 

\subsection{Qualitative Comparison}
Fig. \ref{fig:sota_comp} shows qualitative comparison of current state-of-the-art Cascade R-CNN \cite{cascadercnn} with our F2DNet. It shows that our F2DNet can detect pedestrians even where Cascade R-CNN fails. For results shown in Fig. \ref{fig:sota_comp}, we took models trained on multiple datasets; for Cascade R-CNN we took model weights from Pedestron\cite{pedestron}.

\subsection{Comparison with the State-of-The-Art}
To compare the performance of F2DNet with the current state-of-the-art and top-performing methods, we took models trained on a single dataset without using any extra data except for pre-trained backbones. F2DNet outperforms the existing state-of-the-art in Caltech pedestrian \cite{caltech} and Euro City Persons \cite{eurocitypersons} datasets as well as in heavy occlusion settings of City Persons dataset \cite{citypersons}  with a clear margin and achieves slightly better $MR^{-2}$ in reasonable and small settings of City Persons dataset \cite{citypersons}. 

\begin{figure}[!t]
\centering
\subfloat[][]{\includegraphics[width=0.155\textwidth]{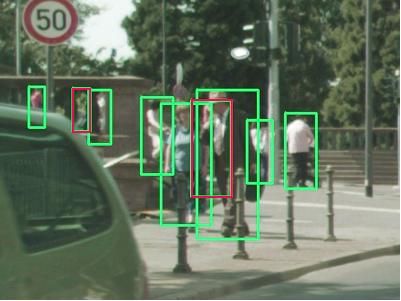} \label{fig:unsup1}}
\subfloat[][]{\includegraphics[width=0.155\textwidth]{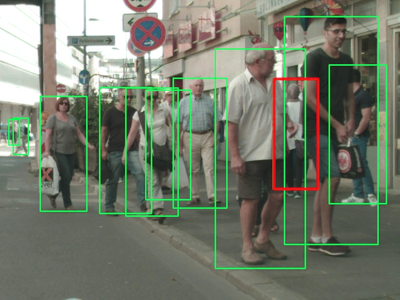} \label{fig:unsup2}}
\subfloat[][]{\includegraphics[width=0.155\textwidth]{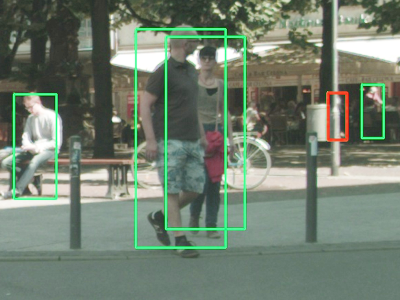} \label{fig:unsup3}}
\vspace{-0.6em}
\subfloat[][]{\includegraphics[width=0.155\textwidth]{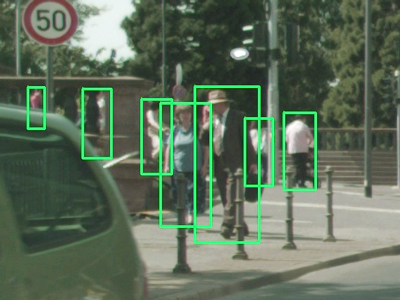} \label{fig:sup1}}
\subfloat[][]{\includegraphics[width=0.155\textwidth]{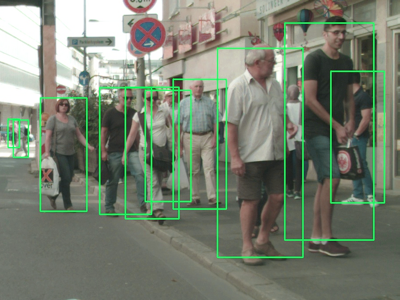} \label{fig:sup2}}
\subfloat[][]{\includegraphics[width=0.155\textwidth]{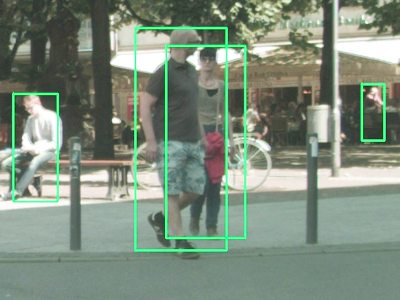} \label{fig:sup3}}
\caption{Results of F2DNet before and after suppression. (a, b, c) Results without suppression; it can be seen that there are few false positives (marked red) in small and heavy occlusion cases. (d, e, f) Results with suppression; the false positives have been successfully suppressed.}
\label{fig:res}
\end{figure}

\subsection{Efficacy of Suppression Head}
 Fig. \ref{fig:res} first row shows that F2DNet without suppression head produces false positives most of which are suppressed by employing fast suppression head as shown in Fig. \ref{fig:res} second row.
However, F2DNet with suppression head performs slightly worse compared to F2DNet without suppression head in heavy occlusion setting of Caltech pedestrian dataset \cite{caltech}. This performance drop can be attributed to the sparseness of the Caltech pedestrian dataset \cite{caltech} compared to other pedestrian datasets with less heavy occlusion samples to train suppression head well. Table \ref{tab:sota} shows the detailed results our experiment.

\section{Cross Dataset Evaluation}
We conduct cross dataset evaluation to test how well F2DNet generalizes to unseen data. We compare the generalizability of F2DNet, with two other models, namely CSP \cite{csp} and Cascade RCNN \cite{cascadercnn}. Both of these models are state of the art in the context of pedestrian detection. We used scores for Cascade RCNN \cite{cascadercnn} and CSP \cite{csp} provided in Pedestron \cite{pedestron}. We train F2DNet only on training sets and conduct tests on the validation set for City Persons \cite{citypersons} and Euro City Persons \cite{eurocitypersons} and on the test set of Caltech Pedestrian dataset \cite{caltech}. F2DNet generalizes better than CSP \cite{csp} and Cascade RCNN \cite{cascadercnn}, in most cases, for City Persons \cite{citypersons} or Euro City Persons \cite{eurocitypersons} (refer to Table \ref{tab:cross_dataset}). However, for Caltech dataset \cite{caltech} F2DNet generalizes slightly worse than other models. F2DNet beats CSP \cite{csp} and Cascade RCNN \cite{cascadercnn} with a large margin when trained on City Persons \cite{citypersons} and tested on Euro City Persons \cite{eurocitypersons}, this shows that F2DNet performs well even when trained on a smaller dataset. Cross dataset evaluation scores can be seen in Table \ref{tab:cross_dataset}.

\begin{table}[!t]
        \renewcommand{\arraystretch}{1.3}
        \caption{Cross dataset evaluation results. Our model is more generaliable compared to CSP and Cascade RCNN is most cases specially when trained on City Persons \cite{citypersons} and tested on Euro City Persons \cite{eurocitypersons}.}
        \label{tab:cross_dataset}
        \centering
        \begin{tabular}{l|c|c|c|c}
             \hline
             \multicolumn{1}{c|} {Method\textbackslash $MR^{-2}$} & Training & Reasonable & Small & Heavy Occ. \\
             \hline
             \hline
             \multicolumn{5}{c}{\textbf{City Persons} \cite{citypersons}}\\
             \hline
             \hline
             CSP  \cite{pedestron, csp} & ECP & 11.5 & 16.6 & 38.2  \\
             \hline
             Cascade RCNN  \cite{pedestron} & ECP & 10.9 & \textbf{11.4} & 40.9  \\
             \hline
             F2DNet (ours) & ECP & \textbf{10.1} & 12.1 &\textbf{ 36.4} \\
             \hline
             \hline
             \multicolumn{5}{c}{\textbf{Caltech} \cite{caltech}} \\
             \hline
             \hline
             F2DNet (ours) & CP & 11.3 & 13.7 & 32.6\\
             \hline
             CSP \cite{pedestron, csp} & CP & 10.1 & 13.3 & 34.4 \\
             \hline
             Cascade R-CNN \cite{pedestron} & CP & \textbf{8.8} & \textbf{9.8} & \textbf{28.8} \\
             \hline
             \hline
             \multicolumn{5}{c}{\textbf{Euro City Persons} \cite{eurocitypersons}} \\
             \hline
             \hline
             CSP \cite{pedestron, csp} & CP &  19.6 & 51.0 & 56.4 \\
             \hline
             Cascade RCNN \cite{pedestron} & CP &  17.4 & 40.5 & 49.3 \\
             \hline
             F2DNet (ours) & CP &  \textbf{11.6} & \textbf{14.7} & \textbf{40.0} \\
             \hline
        \end{tabular}
    \end{table}

\section{Progressive Fine Tuning}
To further improve the performance of F2DNet we perform progressive fine-tuning. We initially train our model on a bigger and diverse dataset and fine-tune it towards the target dataset in cascading manner. For City Persons dataset \cite{citypersons}, we train the model on Euro City Persons \cite{eurocitypersons} and fine-tune on City Persons dataset \cite{citypersons}. For Caltech pedestrian dataset \cite{caltech} we take the fine-tuned model on City Persons dataset \cite{citypersons} and fine-tune it on the Caltech pedestrian dataset \cite{caltech}. Through progressive fine-tuning, we were able to achieve new all times low $MR^{-2}$ in heavy occlusion settings for Caltech Pedestrian \cite{caltech} and City Persons datasets \cite{citypersons} as shown in Table \ref{tab:fine_tune}. For both training and fine-tuning only train sets of respective datasets were used.

\begin{table}[!t]
        \renewcommand{\arraystretch}{1.3}
        \caption{Results of F2DNet trained on multiple datasets in progression fine-tuning fashion.}
        \label{tab:fine_tune}
        \centering
        \begin{tabular}{l|l|c|c|c}
             \hline
             Training & Testing & Reasonable & Small & Heavy Occ. \\
             \hline
             ECP $\rightarrow$ CP & CP & \textbf{7.80} & \textbf{9.43} & \textbf{26.23} \\
             \hline
             ECP $\rightarrow$ CP $\rightarrow$ Caltech & Caltech & \textbf{1.71} & \textbf{2.10} & \textbf{20.42}\\
             \hline
        \end{tabular}
    \end{table}

%% file: conclusion.tex
\section{Conclusion}
Two-stage detectors perform well in pedestrian detection. However, the region proposal network-based two-stage detectors are inefficient as the region proposal networks predict weak proposals, dependent on further refinement. We replaced the region proposal network with a per-pixel center and scale regression based focal detection network to produce high-quality, standalone detection candidates except for few false positives in small and occluded settings. We pass these strong detection candidates through a light yet fast suppression network, which with barely noticeable computational cost further refines the detections to produce promising results. Our model beats the state-of-the-art in most visibility and height settings while being on par in rest. Also, by using Euro City Persons \cite{eurocitypersons} and City Persons \cite{citypersons} datasets as extra training data, our model achieves the lowest $MR^{-2}$ in a heavy occluded setting, in a multi-dataset setup.